%% file: main.tex
\pdfoutput=1
\documentclass[11pt]{article}
\usepackage{acl}
\usepackage{times}
\usepackage{latexsym}
\usepackage[T1]{fontenc}
\usepackage[utf8]{inputenc}
\usepackage{microtype}
\usepackage{inconsolata}
\usepackage{graphicx}
\usepackage{multirow}
\usepackage{diagbox}
\usepackage{booktabs}
\usepackage{amsmath, amssymb, bm}
\usepackage{algorithm}
\usepackage{algorithmic}
\floatname{algorithm}{Algorithm}

\title{Review-LLM: Harnessing Large Language Models \\ for Personalized Review Generation}

\author{
Qiyao Peng\textsuperscript{1}, Hongtao Liu\textsuperscript{2},
Hongyan Xu\textsuperscript{3}, Qing Yang\textsuperscript{2}, Minglai Shao\textsuperscript{1},
Wenjun Wang\textsuperscript{3} \\
\textsuperscript{1} School of New Media Communication, Tianjin University, China \\
\textsuperscript{2} Du Xiaoman Financial, Beijing, China \\
\textsuperscript{3} College of Intelligence and Computing, Tianjin University, China \\
\textsuperscript{1}\texttt{\{qypeng, shaoml\}@tju.edu.cn} \\
\textsuperscript{2}\texttt{\{liuhongtao01, yangqing\}@duxiaoman.com} \\
\textsuperscript{3}\texttt{\{hongyanxu, wjwang\}@tju.edu.cn}
}

\begin{document}
\maketitle

\input{0-abstract}
\input{1-introduction}
\input{2-methods}
\input{3-experiments}
\input{4-conclusion}

\clearpage

\input{5-limitation}

\bibliography{ref}

\end{document}

%% file: 0-abstract.tex
\begin{abstract}
Product review generation is an important task in recommender systems, which could provide explanation and persuasiveness for the recommendation.
Recently, Large Language Models (LLMs, e.g., ChatGPT) have shown superior text modeling and generating ability, which could be applied in review generation.
However, directly applying the LLMs for generating reviews might be troubled by the ``polite'' phenomenon of the LLMs and could not generate personalized reviews (e.g., negative reviews).
In this paper, we propose Review-LLM that customizes LLMs for personalized review generation.
Firstly, we construct the prompt input by aggregating user historical behaviors, which include corresponding item titles and reviews. 
This enables the LLMs to capture user interest features and review writing style. 
Secondly, we incorporate ratings as indicators of satisfaction into the prompt, which could further improve the model's understanding of user preferences and the sentiment tendency control of generated reviews.
Finally, we feed the prompt text into LLMs, and use Supervised Fine-Tuning (SFT) to make the model generate personalized reviews for the given user and target item.
Experimental results on the real-world dataset show that our fine-tuned model could achieve better review generation performance than existing close-source LLMs.
\end{abstract}

%% file: 1-introduction.tex
\section{Introduction}

Online e-commerce platforms (e.g., Amazon.com) usually offer users opportunities to share reviews for items they have purchased~\cite{sun2020dual}.
These reviews typically contain rich user preference information and detailed item attributes~\cite{mcauley2013hidden}, which can inform users about the item and improve recommendation accuracy.
However, many users only provide a rating for the item but no review after purchasing the item.
Therefore, review generation task has attracted more attentions~\cite{lu2018like}.

Most existing methods are based on the encoder-decoder neural network framework~\cite{li2019generating,li2020knowledge,kim2020retrieval}.
Earlier methods utilize discrete attribute information about users and items to generate reviews~\cite{tang2016context,dong2017learning,ni2017estimating,zang2017towards}.
For example, Tang et al.~\cite{tang2016context} utilize user/item IDs, and rating as input information, and use the RNN-based decoder for generating reviews.
Recent works consider using the text information to help generating reviews, such as item titles, and historical reviews of users/items, etc~\cite{ni2018personalized,li2019towards}.
Ni et al.~\cite{ni2018personalized} propose ExpansionNet,
which also integrates phrase information from item titles and review summaries into the encoder for generating reviews.
Li et al.~\cite{li2019towards} propose a RevGAN model to generate controllable and personalized reviews from item descriptions and sentiment labels.

Recently, owing to the strong reasoning and learning capabilities exhibited by Large Language Models (LLMs)~\cite{achiam2023gpt,touvron2023llama}, many researchers are extending LLMs applications in other domains, such as Recommender Systems (RS)~\cite{xu2024prompting}.
Motivated by this, in this paper, we want to preliminary explore how to extend the LLMs (e.g., Llama-3) to the review generation.
Compared with other traditional generation tasks (such as poem generation), applying LLMs for the review generation in the e-commerce platforms is more challenging due to the lack of personalized information.
First, most existing large language models are usually pre-trained at the corpus-level and might not capture the review style and habits of the users.
This might cause the generated review to be inconsistent with user's previous reviews.
Second, users are dissatisfied with many items and the corresponding reviews should be negative. 
However, the generated text by the LLMs is usually ``polite''~\cite{touvron2023llama}, which might lead to the model generating positive reviews for the user's dissatisfaction.

Hence, in this paper, we design a framework (Review-LLM) for harnessing the LLMs to generate personalized reviews.
Specifically, we re-construct the model input via aggregating the user behavior sequence, including the item titles and corresponding reviews.
In this way, the model could learn user interest features and review writing styles from semantically rich text information.
Furthermore, the user's rating of the item can be used to indicate the user's satisfaction with the item. 
We integrate this information into the prompt input accordingly.
In this way, the large language model can better perceive whether users like different items, and may prevent the model from generating more ``polite'' reviews.
Finally, we feed the input prompt text into the LLMs (Llama-3), which is subsequently fine-tuned using Supervised Fine-Tuning (SFT) to output the review for target items.
For experiments, we design different difficulty levels review generation testing dataset to verify the effectiveness of different models.

%% file: 2-methods.tex
\section{Method}

\begin{figure*}
\centering
\includegraphics[width=0.96\textwidth]{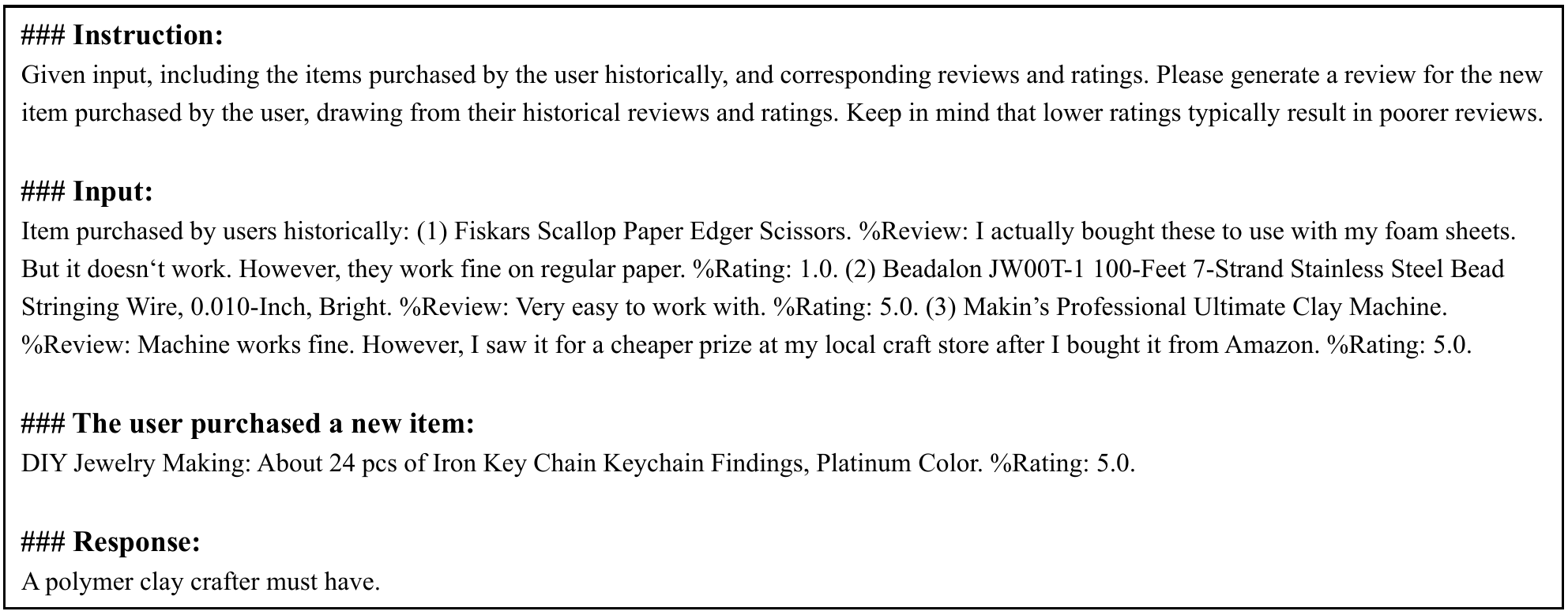}
\caption{An example of input prompt for Review-LLM.}
\label{fig:ft} 
\end{figure*}

\subsection{Problem Formulation}

Given the user $u$, item $v$, rating $r$, and user's historical interaction, review generation aims to automatically generate personalized reviews for the user $u$ towards the target item $v$.
Especially, the user's historical interaction is a sequence of items that the user purchased, which can be denoted as  $H^{u} = \{v_1, v_2, \cdots, v_h\}$, where $h$ is the number of items.
And corresponding rating score sequence $R^{u} = \{r_1, r_2, \cdots, r_h\}$, where $h$ is the number of ratings.
The $i$-th item title and corresponding review are denoted as: $T^{u}_i = \{w_1, w_2, \cdots, w_N \}$ and $E^{u}_i = \{ w_1, w_2, \cdots, w_M \}$ respectively, where $N$ and $M$ are their lengths.
We denote the generated review as $\hat{Y} = \{w_1, w_2, \cdots, w_L \}$ and $L$ is the length; the reference review is denoted as $Y = \{w_1, w_2, \cdots, w_{L'} \}$ and $L'$ is the length.

\subsection{Review-LLM}

In this section, we introduce Review-LLM for generating reviews. 
The key is to enhance the LLMs to learn more personalized user interest features and review writing styles based on the histories.
Specifically, we propose to construct a prompt text for training the LLM-based model using a supervised fine-tuning approach.
As shown in Figure~\ref{fig:ft}, the prompt text composes of the following parts:

\textbf{1) Generation Instruction}: Its role is to instruct the LLMs to consider both the user's preference and historical behaviors to complete the generation task. The generation task is structured as an output of the review for the target item;
\textbf{2) Input}: This contains the items the user has interacted with, including the item title, review, and rating;
\textbf{3) The user purchased a new item}: This contains the target item title and the corresponding rating;
\textbf{4) Response}: This is the generated review for the target item.

Then, we use the following SFT training loss to train the LLM-based review generation model:
\begin{equation}
    \mathcal{L}_{sft} = - \sum_{i=1}^{L}logp(w_i|w_{<i}),
\end{equation}
where $w_i$ is the $i$-th word in the generated review and $L$ is the length of that. 
The probability $p(w_i|w_{<i})$ is calculated by the LLM model following the next-token prediction paradigm.
During the training process, we utilize the Low-Rank Adaptation (LoRA)~\cite{hu2021lora} for Parameter-Efficient Fine-Tuning (PEFT), which can greatly reduce the number of trainable parameters.

During inference, we remove the review of the target item in the \textbf{4) Response}.
Then we input this modified prompt into the large language model to generate the review for the target item.

%% file: 3-experiments.tex
\section{Experiments}

\subsection{Experimental Setting}

In this paper, we select five open-source 5-core recommendation datasets from Amazon dataset~\footnote{\url{https://cseweb.ucsd.edu/~jmcauley/datasets/amazon_v2/}}, including ``Arts, Crafts and Sewing'', ``Office items'', ``Musical Instruments'', ``Toys and Games'' and ``Video Games''.
We only remain users with more than 10 historical interactions and less than 30 historical interactions.
We timely sort user interactions, then employ the last review as the reference review, and treat others as historical interactions.
Then, we randomly select 1000 samples from each dataset as the training set and 200 samples as simple evaluation data from the remaining data.
Furthermore, we select 200 negative reviews from each dataset as hard evaluation data to test the model's ability to generate negative reviews.

We conduct experiments using a cluster composed of 4*A800 80GB GPUs. 
We select Llama-3-8b~\footnote{\url{https://llama.meta.com/llama3/}} as the base model. 
And, we conduct the SFT training based on PyTorch and PEFT library~\cite{peft} and use the LoRA~\cite{hu2021lora} with a rank equal to 8. 
In addition, we use the Adam optimizer with learning rate of 5e-6 and batch size of 1 for SFT, and we set gradient accumulation steps as 2.
We conduct each experiment independently and repeat it 5 times, and report the average results.

\subsection{Baselines and Evaluation Metrics} 

We compare Review-LLM with: (i) closed-source models such as GPT-3.5-Turbo, GPT-4o~\cite{achiam2023gpt}; (ii) open-source models such as, Llama-3-8b~\cite{touvron2023llama}.

To evaluate the performance of different models, we select ROUGE-1/L~\cite{lin-2004-rouge} and BERT~\cite{kenton2019bert} similar score (BertScore) as evaluation metrics.
ROUGE-n measures the n-gram similarity while BertScore measures the semantic similarity in the embedding space between the generated reviews and the reference reviews.
We use the sentence transformers~\cite{reimers-2019-sentence-bert} to compute the BertScore.
Besides, we conduct a human evaluation experiment to test whether the generated reviews are semantically consistent with the reference reviews.

\begin{table}
\centering
\caption{Simple evaluation. w/ rating means the prompt contains ratings and w/o rating is vice.}
\resizebox{0.42\textwidth}{!}{
\begin{tabular}{c|c|c|c}
\toprule[1.5pt]
\diagbox{Method}{Metric} & ROUGE-1 & ROUGE-L & BertScore (mean)  \\
\midrule
GPT-3.5-turbo (w/ rating) & 15.99 & 9.84 & 41.52  \\
GPT-3.5-turbo (w/o rating) & 16.00 & 9.81 & 41.37 \\
GPT-4o (w/ rating) & 12.80 & 8.47 & 40.12 \\
GPT-4o (w/o rating) & 15.41 & 11.22 & 41.73 \\
Llama-3-8b (w/ rating) & 12.23 & 8.23 & 31.30 \\
Llama-3-8b (w/o rating) & 13.82 & 9.59 & 30.46 \\
\textbf{Review-LLM (w/ rating)} & \textbf{31.15} & \textbf{26.88} & \textbf{49.52} \\
Review-LLM (w/o rating) & 30.47 & 26.38 & 48.56 \\
\bottomrule[1.5pt]
\end{tabular}
}
\label{tab:result}
\end{table}

\subsection{Overall Performance}

Table~\ref{tab:result} compares the performance of our method with several baselines and ablations. 
It is noted that the GPT-3.5-Turbo and GPT-4o are always better than Llama-3-8b, the reason is that the GPT-series models have a larger number of parameters and are pre-trained on massive data, which could learn more general knowledge.
Besides, we find that some baselines without ratings perform better than with ratings, while our fine-tuning method is the opposite. 
We argue that this is because the user rating information is further pre-trained in our method while baselines not.
Overall, our method Review-LLM outperforms all methods (including GPT-3.5-Turbo and GPT-4o) across all metrics, demonstrating the effectiveness of using the item title, review, and rating to personalized fine-tune.

\subsection{Negative Review Performance}

\begin{table}
\centering
\caption{Hard evaluation.  w/ rating means the prompt contains ratings and w/o rating is vice.}
\resizebox{0.42\textwidth}{!}{
\begin{tabular}{c|c|c|c}
\toprule[1.5pt]
\diagbox{Method}{Metric} & ROUGE-1 & ROUGE-L & BertScore (mean)  \\
\midrule
GPT-3.5-turbo (w/ rating) & 17.62 & 10.70 & 37.45  \\
GPT-3.5-turbo (w/o rating) & 16.07 & 9.89 & 37.25 \\
GPT-4o (w/ rating) & 16.66 & 9.86 & 39.21 \\
GPT-4o (w/o rating) & 14.51 & 8.73 & 38.64 \\
Llama-3-8b (w/ rating) & 13.47 & 8.05 & 28.38 \\
Llama-3-8b (w/o rating) & 13.11 & 7.89 & 26.96 \\
\textbf{Review-LLM (w/ rating)} & \textbf{21.93} & \textbf{16.63} & \textbf{39.35} \\
Review-LLM (w/o rating) & 17.82 & 13.50 & 35.89 \\
\bottomrule[1.5pt]
\end{tabular}
}
\label{tab:badresult}
\end{table}

In our method, we employ user rating information to strengthen the model’s understanding of user preferences for different items to achieve more personalized review generation.
In this part, we test the performance of the model on the constructed hard testing set. 
The different model performance is shown in Table~\ref{tab:badresult}.
From the results, we can find that all model performance has decreased compared with Table~\ref{tab:result}.
In particular, using Llama3-8b for inference directly, BertScore is reduced to $26.96$.
We argue that this is because the LLMs might be polite, resulting in insufficient negative information captured during generating reviews.
Besides, methods with ratings outperform methods without ratings on semantic similarity, especially Review-LLM, which further confirms the necessity of fusing the rating information for personalized review generation.

\subsection{Human Evaluation}

\begin{figure}
\centering
\includegraphics[width=0.46\textwidth]{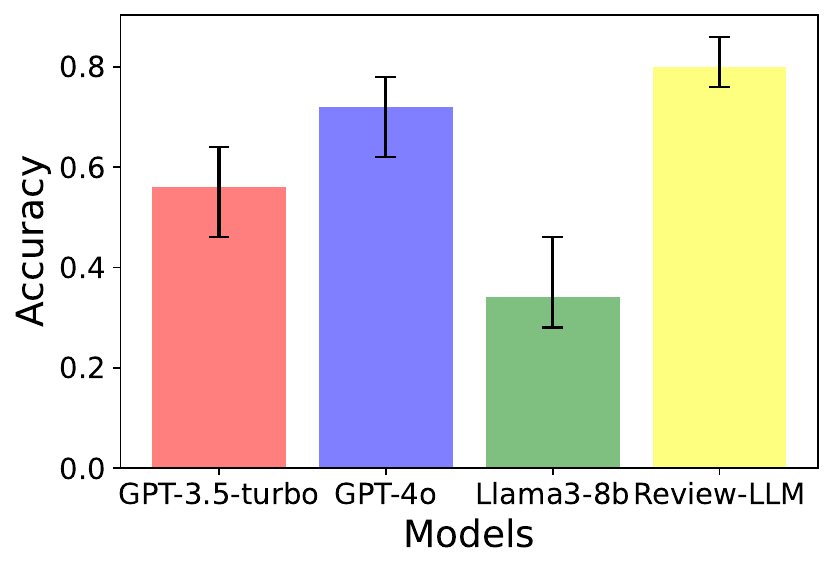}
\caption{Human Evaluation. The bar is the mean of the model performance, and the error bar represents the max and min accuracy of the model.}
\label{fig:humanresult} 
\end{figure}

In this part, we conduct the human evaluation to test the model performance of review generation.
Considering that the generated texts with rating information usually have higher semantic similarity than those without, we only compare the models with rating information here.
We randomly select 100 reference reviews and generated reviews from the simple testing set, and hire 10 Ph.D. students who are familiar with review/text generation to evaluate the similarity between generated reviews and reference reviews.
If the reference review is semantically similar to the generated reviews, it is marked as $1$, otherwise it is marked as $0$.
Figure~\ref{fig:humanresult} shows the percentage of generated reviews marked as $1$.
From the results, we can see that the designed fine-tuning data and framework could improve the quality of generated reviews and increase their semantic similarity to the reference reviews.

\subsection{Case Study}

\begin{figure}
\centering
\includegraphics[width=0.44\textwidth]{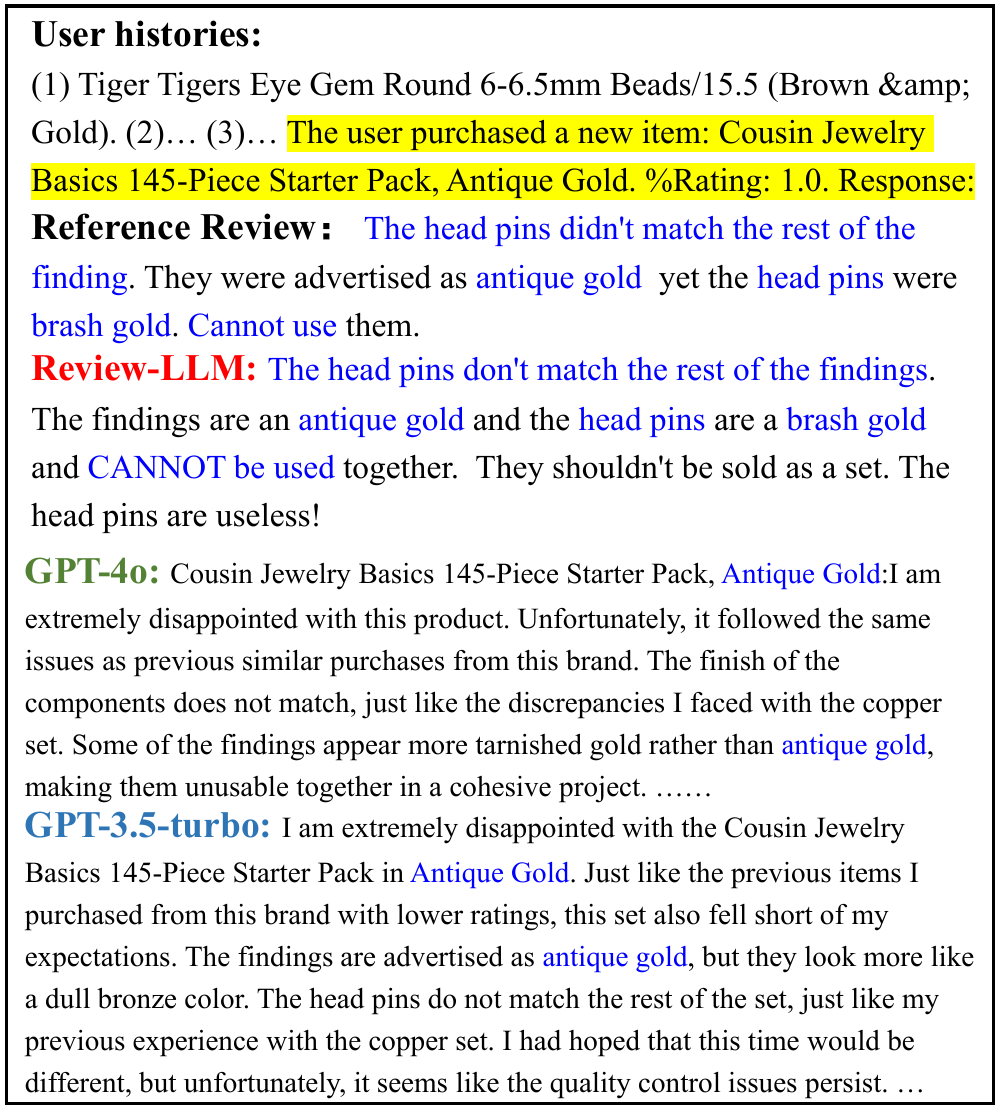}
\caption{Case Study.}
\label{fig:casestudy} 
\end{figure}

To demonstrate the effect of Review-LLM on generating reviews more intuitively, we select the generated reviews (from Review-LLM, GPT-4o, GPT-3.5-Turbo) and the real review for visualization shown in Figure~\ref{fig:casestudy}.
First, we can find the review generated by our model is semantically similar to the real review and brief.
In contrast, reviews derived by GPT-3.5-Turbo/GPT-4o are too long and may not be suitable for e-commerce platforms.
Second, the generated review of Review-LLM better reflects review writing styles and user sentiment towards the item (we marked those in blue font).
This demonstrates that our model could generate high-quality personalized reviews effectively by unifying rich user information with LLMs.

%% file: 4-conclusion.tex
\section{Conclusion}

This paper presents a framework that leverages Large Language Models (LLMs) for personalized review generation in recommender systems. 
By aggregating user historical behaviors, including item titles, reviews, and ratings, we construct a comprehensive input prompt to capture user preferences and review writing style. 
In this way, the model could mitigate the generation of overly polite reviews.
Then, we utilize the low-rank adaptation for parameter-efficient fine-tuning, enabling the LLMs to generate reviews for candidate items through supervised fine-tuning.
Experimental results show that our fine-tuning method outperforms GPT-3.5-Turbo and GPT-4o in review generation.

%% file: 5-limitation.tex
\section{Limitation}

(1) Different individuals may focus on different aspects of a product, such as price, quality, appearance, or durability. While the proposed framework leverages user historical behaviors to capture comprehensive user interest features, it may not fully capture the diversity of individual preferences.
(2) The framework primarily focuses on capturing user preferences from historical behaviors without considering the dynamics of user interactions over time. User preferences and writing styles can evolve, and incorporating temporal dynamics could potentially improve the accuracy and personalization of review generation.